\documentclass{article}

\usepackage{microtype}
\usepackage{graphicx}
\usepackage{subfigure}
\usepackage{booktabs} 

\usepackage{hyperref}

\usepackage[accepted]{tccmlicml2021}

\begin{document}

\twocolumn[
\icmltitle{Self-supervised Contrastive Learning for Irrigation Detection \\
in Satellite Imagery}



\icmlsetsymbol{equal}{*}

\begin{icmlauthorlist}
\icmlauthor{Chitra S. Agastya}{equal,berkeley,ibm}
\icmlauthor{Sirak Ghebremusse}{equal,berkeley}
\icmlauthor{Ian Anderson}{equal,berkeley}
\icmlauthor{Colorado Reed}{berkeley}
\icmlauthor{Hossein Vahabi}{berkeley}
\icmlauthor{Alberto Todeschini}{berkeley}
\end{icmlauthorlist}

\icmlaffiliation{berkeley}{School of Information, University of California, Berkeley, USA}
\icmlaffiliation{ibm}{IBM Chief Analytics Office, Armonk, New York, USA}

\icmlcorrespondingauthor{Chitra S. Agastya}{chitra.agastya@berkeley.edu, Chitra.S.Agastya@ibm.com}
\icmlcorrespondingauthor{Sirak Ghebremusse}{sirakzg@berkeley.edu}
\icmlcorrespondingauthor{Ian Anderson}{imander@berkeley.edu}

\icmlkeywords{SimCLR, Sentinel2, ICML, irrigation, deep learning}

\vskip 0.3in
]



\printAffiliationsAndNotice{\icmlEqualContribution} 

\begin{abstract}
Climate change has caused reductions in river runoffs and aquifer recharge resulting in an increasingly unsustainable crop water demand from reduced freshwater availability.
Achieving food security while deploying water in a sustainable manner will continue to be a major challenge necessitating careful monitoring and tracking of agricultural water usage. 
Historically, monitoring water usage has been a slow and expensive manual process with many imperfections and abuses. 
Machine learning and remote sensing developments have increased the ability to automatically monitor irrigation patterns, but existing techniques often require curated and labelled irrigation data, which are expensive and time consuming to obtain and may not exist for impactful areas such as developing countries. 
In this paper, we explore an end-to-end real world application of irrigation detection with uncurated and unlabeled satellite imagery. We apply state-of-the-art self-supervised deep learning techniques to optical remote sensing data, and find that we are able to detect irrigation with up to nine times better precision, 90\% better recall and 40\% more generalization ability than the traditional supervised learning methods.
\end{abstract}

\section{Introduction}

Water is essential for global food production and security, agriculture being the largest consumer of freshwater globally \cite{wwap}. While we need water to grow crops to address the growing demand for food, excessive use of water has negatively impacted ecosystem and adversely affected crop production and livelihood. Ground truth data on agricultural water demand, traditionally gathered from local surveys, are often inadequate and do not capture the temporal signature and dynamics of irrigation patterns \cite{Lukas20}. Irrigated land is difficult to track in underdeveloped countries due to lack of infrastructure and government policies around water usage.

Detecting irrigation is critical to understand water usage and promote better water management. Such data will potentially enable the study of climate change impact on agricultural water sources, monitor water usage, help detect water theft and illegal agriculture and inform policy decisions and regulations related to water compliance and management. This is a particularly hard problem to solve due to the lack of curated and labelled data available that are centered around irrigation systems. BigEarthNet-S2, for example, is a small fraction of the 26 TB of data produced daily by the European Space Agency, and required several million dollar grants to curate and label [\citenum{tu_berlin}].

The small sample of labelled datasets currently available are from developed countries and supervised learning from these may not generalize to developing and under developed countries. In this report we investigate the problem of irrigation detection using self-supervised learning on high resolution multi-spectral satellite images that were collected for monitoring of land, vegetation, soil and water cover. We research whether pre-training with contrastive self-supervised learning from uncurated satellite images of one geography, followed by fine-tuning with a small fraction of publicly available labelled data from a different geography, has the generalization ability to both detect irrigated land from the first geography and also from entirely different geographical regions.

\section{Methodology}
We use the following machine learning pipeline to build an irrigation detection system: (1) We pre-train a deep neural network using a large amount of unlabeled data from a geographic area of interest, i.e.~where we would like to do irrigation detection. (2) We fine-tune the network using publicly available labeled satellite image datasets that contain irrigation labels. (3) We apply the irrigation detection system to the geographic area of interest used for pre-training in (1.). For (1.) we use the SimCLR framework\cite{chen2020simple, chen2020big} because it consistently outperformed previous methods for self-supervised and semi-supervised learning on ImageNet, without requiring specialized architectures.

\section{Data}

The data for our research are Sentinel2 surface reflectance images collected from three sources. We use Sentinel2 archive of level2A images of the Central Valley region of California as our unlabelled source for self-supervised pretraining. Central Valley is one of the most productive agricultural regions in the world~[\citenum{amnh}]. The Central Valley is home to more than 7 million acres of permanently irrigated land and houses many major cities such as Sacramento and Fresno, providing a diverse mix of irrigated and non-irrigated land for our study [\citenum{usgs}]. 

BigEarthNet-S2 \cite{Sumbul_2019}, a large scale benchmark archive for remote sensing image understanding from ten European countries, is used in our experiments for supervised training and evaluation. For evaluating model generalization, we use Sentinel2 images of ground truth coordinates spanning six different countries from croplands.org\cite{samerica, seasia, sasia, africa}.


\section {Experiments}
We established our baseline using traditional supervised learning approaches on the BigEarthNet-S2 data. Our target label is a binary variable indicating whether the land in the image is permanently irrigated or not. We conducted several experiments by varying different parameters: (i) size of the CNN architecture, (ii) training from scratch versus using pretrained ImageNet weights (iii) size of the training data. We use varying training data sizes of  190, 570, 1902, 4756, 9515 and 19024 records balanced between the two target classes. We also refer to these as data split percentages of 1, 3, 10, 25, 50 and 100 respectively. 

\subsection{SimCLR-S2 Model}

We call our adaptation of the SimCLR framework to Sentinel2 images, \textit{SimCLR-S2}. While the original SimCLR implementation uses only the 3 visible RGB channels, we have access to 10 channels (bands) in our data. Some of these, like infrared bands, provide information about the presence of moisture due to vegetation. We therefore consider using all 10 bands throughout training and inference.

The SimCLR-S2 framework has three stages. The first stage takes an unlabeled dataset and creates a mini batch of images. For each image in the mini batch, we perform two image augmentations, picked at random from a list of transformation pairs, and compute the contrastive loss between any two images in the mini batch with an objective to minimize the contrastive loss across the entire dataset. In the second stage, the pretrained self-supervised model is fine tuned with a small labelled dataset for the task of identifying irrigated land versus non-irrigated land. In the third stage, the fine tuned model is then used as a teacher in a knowledge distillation process to teach labels for an unlabeled dataset on a student network of the same size or smaller. The student network bootstraps from the teacher network, learning nuances from the teacher, resulting in performance equal to or better than the teacher model, even on smaller networks.

Since BigEarthNet is our only source of annotated data, we use 3\% of its labeled data for evaluation. We performed several experiments with variations in: data used for unsupervised pretraining; data used in supervised fine-tuning; and CNN architectures used in self-training distillation stages. We then compare the results for these experiments with the supervised baseline results. 

\begin{figure}
    \vskip 0.1in
    \begin{center}
    \centerline{\includegraphics[width=\columnwidth]{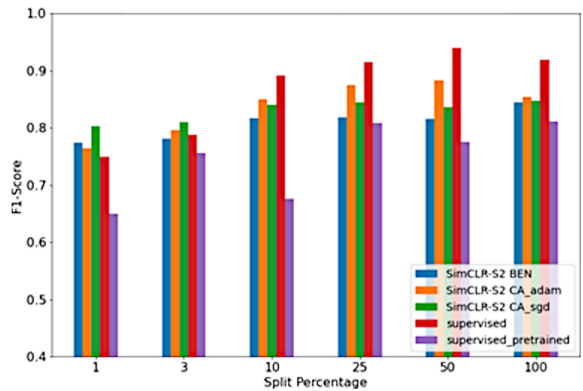}}
    \caption{A comparison of stage-2 fine-tune results of SimCLR-S2 with supervised baseline for the different data sizes (given by split percentage with 1\% having 190 training records). SimCLR-S2 models (i) consistently outperform supervised baseline scores pretrained with ImageNet weights and (ii) outperform baselines scores when trained from scratch on smaller sizes of annotated data (1\% and 3\% splits)}
    \label{fig:finetune_results}
    \end{center}
    \vskip -0.1in
\end{figure}

\subsubsection{Hyper-parameter Tuning}
We evaluated training SimCLR-S2 with the SGD and Adam optimizers. With SGD optimization with cosine decay, the contrastive loss did not decrease throughout training, while with the Adam optimizer, we observed decreasing contrastive loss curves. As a result, we selected the Adam optimizer for our pipeline. A learning rate of 0.0005 had the best performance with the Adam optimizer.

\section{Evaluation Methods and Results}
\label{results}
We conducted two types of evaluations to compare the performance of SimCLR-S2 model to that of the supervised baseline model. We evaluated the F-1 scores of predictions from both models on held-out data from BigEarthNet-S2 dataset. This test data was not a part of the training process. However, because both the test and training data are from the same region i.e. Europe, we evaluate the generalization ability of the self-supervised vs supervised models by comparing: (i) the precision scores on an unlabelled set of Sentinel2 images (ii) the recall scores on Sentinel2 images for known irrigated land co-ordinates from different countries.

\subsection {Results on BigEarthNet-S2 hold out data }

The results from our fine-tuning experiments showed that the SimCLR-S2 model outperformed the supervised baseline on smaller data sizes. This is evidenced in figure \ref{fig:finetune_results}, where for 1\% of data splits, the F-1 scores were better or on par with that of the supervised learning. Our SimCLR-S2 model consistently outperformed the supervised model that was pretrained with ImageNet weights.

\begin{figure}[!htb]
    \vskip 0.1in
    \begin{center}
    \centerline{\includegraphics[width=0.5\textwidth]{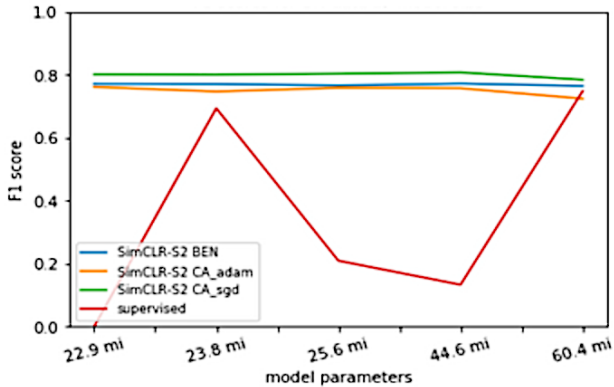}}
    \caption{A comparison of stage-3 distillation results of SimCLR-S2 with supervised baseline for different architectures (in number of million parameters) for training data size of 190 records. SimCLR-S2 models outperform supervised baseline scores across architectures.}
    \label{fig:acc1}
    \end{center}
    \vskip -0.1in
\end{figure}

\begin{figure}[!hb]
    \vskip 0.1in
    \begin{center}
    \centerline{\includegraphics[width=0.5\textwidth]{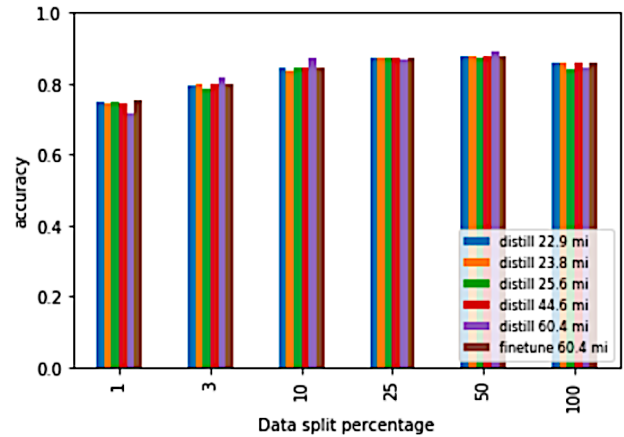}}
    \caption{A comparison of accuracy with and without distillation across different data sizes for different architectures (in number of million parameters).Most distilled models perform on par or better that a simple finetuned model (i.e. without distillation) across data sizes even on smaller architectures.}
    \label{fig:distill}
    \end{center}
    \vskip -0.1in
\end{figure}

We took each ResNet152 model fine-tuned across the various data splits and performed distillation learning on five model architectures. We found that even the smallest model Xception, with 22.9 million parameters,  saw an improved F-1 score for data splits from 3\% and above, and with larger models accuracy over the fine-tuned teacher model increased across the board. We suspect that the freezing of the convolutional layers during the fine-tuning process resulted in their poorer performance compared to the student models, which had all layers enabled for distill learning. Distillation scores were better than those of supervised baselines for smaller data sizes on many architectures. As evidenced by Figure \ref{fig:acc1}, for the 1\% data split SimCLR-S2 outperformed the supervised baseline across the board. 

\subsection{Results on generalizability tests}
We ran two case studies to check for generalizability of our SimCLR-S2 model. In one, we tested for model precision on an unlabelled test dataset and in the second one, we tested for model recall on a labelled test dataset. To demonstrate that the model generalizes well, we used data from diverse geographies different from the geography that the model was trained on. 

\subsubsection{Precision}
For precision, we evaluated Sentinel2 images of unlabelled data from California, USA on our SimCLR-S2 fine-tuned model. We took the top 100 images that our model predicted as irrigated with at least 99\% confidence. We then visually inspected these images to score the precision. We crowdsourced the visual inspection using Amazon Mechanical Turk and did an additional visual verification of the scores to ensure that the survey responses aligned with our expectations. 

To compare these scores with our supervised baseline, we performed similar evaluation and visual inspection with our supervised baseline model. An interesting observation we made was that prediction confidence of the supervised baseline was far lower than that of the SimCLR-S2 model. While we could determine the top 100 predictions with a minimum confidence threshold of 99\% with SimCLR-S2 models, with the supervised models we had to drop the threshold to 50\% to obtain the top 100 predictions. \textit{Table \ref{tab:precision}} shows the median precision scores. SimCLR-S2 model consistently outperformed supervised baseline in all the data splits indicating that our generalized well on unseen data (see Figure 5 in the supplementary material).

\begin{table}[!t]
\caption{A comparison of precision scores from SimCLR-S2 and supervised baseline on unseen data from different geography.}
\label{tab:precision}
\vskip 0.1in
\begin{center}
 \begin{tabular}{||p{0.3\columnwidth}|p{0.25\columnwidth}|p{0.25\columnwidth}||} 
 \hline
 Training data size (num records) & Precision (SimCLR-S2) & Precision (supervised) \\ [0.5ex] 
 \hline\hline
 190 & \textbf{0.99} & 0.11 \\ 
 \hline
 570 & \textbf{1.00} & 0.2 \\
 \hline
 1902 & \textbf{0.99} & 0.36 \\
 \hline
 4756 & \textbf{0.98} & 0.95 \\
 \hline
 9515 & \textbf{1.00} & 0.78 \\ 
 \hline
 19024 & \textbf{1.00} & 0.47 \\ 
 \hline
\end{tabular}
\end{center}
\vskip -0.1in
\end{table}

\begin{table}[!b]
\caption{A comparison of recall scores on SimCLR-S2 and supervised baseline for irrigated cropland from diverse geographies}
\label{tab:recall}
\vskip 0.1in
\begin{center}
\begin{tabular}{||p{0.17\columnwidth}|p{0.17\columnwidth}|p{0.27\columnwidth}|p{0.18\columnwidth}||} 
 \hline
 Country & Training data (num records) & Recall (SimCLR-S2) & Recall (supervised) \\ [0.5ex] 
 \hline\hline
 Brazil & 190 & \textbf{0.75} & 0.5 \\ 
 \hline
 India & 190 & \textbf{0.9} & 0.67 \\
 \hline
 Indonesia & 570 & \textbf{0.76} & 0.07 \\
 \hline
 Tunisia &  570 & 0.78 & \textbf{0.91} \\ 
 \hline
 Vietnam, Myanmar & 190 & \textbf{0.9} & 0.00 \\ 
 \hline
\end{tabular}
\end{center}
\vskip -0.1in
\end{table}

\subsubsection{Recall}
For recall, we chose to evaluate our models' prediction against global crowdsourced ground truth data from croplands.org. We sampled 100 irrigated cropland coordinates spanning 6 different geographies: Brazil, India, Indonesia, Myanmar, Tunisia and Vietnam. Since the ground truth data is not verified at source, before processing, we visually inspected the raw images to ensure that they looked like irrigated cropland. We then compared our model predictions to the original target label to compute recall. 

Our results demonstrated that SimCLR-S2 model generalized well across geographies. A model  trained with just a fraction\footnote{BigEarthNet-S2 has over 575,000 images, 13,589 of which are irrigated cropland. With SimCLR-S2, the smallest training data only uses 0.033\% of overall images and 0.69\% of irrigated images for supervised fine-tuning.} of the overall annotated data for the task at hand, was able to outperform the supervised baseline model, in most cases, as evidenced by the recall scores in \textit{table \ref{tab:recall}}.

\section{Conclusion}

SimCLR-S2 is successful in detecting irrigation in multispectral images, as evidenced by our results. Through our experiments, we showed that SimCLR-S2 can be used on satellite imagery consisting of several more channels than in typical computer vision applications; and performs better than supervised models in certain scenarios. Our results demonstrated that the SimCLR paradigm consistently outperformed supervised learning using significantly smaller sizes of labelled data, and that these improvements can be distilled into smaller models with fewer parameters. We also show that the SimCLR-S2 model generalizes well across diverse geographies. Where manual annotation is expensive and time consuming, we showed that real-world, uncurated classification tasks of satellite images benefit from contrastive self-supervised  learning to perform image classification using a significantly smaller fraction of labelled images, while still achieving better results than supervised learning methods.

We identify a few areas for future work. While we performed all training with a fixed number of training epochs (50), previous works done by Chen et al\yrcite{chen2020big} indicated they trained on as many as 800 epochs for unsupervised pretraining. We also acknowledge that during our investigation into the most effective augmentations to pair for training, a larger subset of the training data, and more epochs, may have been beneficial. The SimCLR-S2 model, with these improvements,  could pave the path for a new state of the art in detecting and tracking global irrigation data using high resolution satellite imagery.


\nocite{oord2019}
\nocite{bachman2019}
\nocite{henaff2020}
\nocite{he2020}
\nocite{khosla2021}
\nocite{goyal2021}
\nocite{reed2021self}
\nocite{reed2021selfaugment}

\bibliographystyle{icml2021}
\bibliography{main}

\end{document}